\documentclass[letterpaper, 10 pt, conference]{ieeeconf}  

\IEEEoverridecommandlockouts                              

\usepackage{times} 
\usepackage{amsmath} 
\usepackage{amssymb}  

\usepackage{graphicx}
\graphicspath{ {images/} {data/} }

\usepackage{verbatim}
\usepackage{array}
\newcolumntype{C}[1]{>{\centering\arraybackslash}p{#1}}
\usepackage{bm}
\newcommand{\bsigma}{\bm{\sigma}}
\DeclareMathOperator*{\argmax}{argmax}
\usepackage{mathtools}
\DeclarePairedDelimiter{\abs}{\lvert}{\rvert}
\usepackage{siunitx}
\usepackage{multirow}
\usepackage{tabularx}
\usepackage{makecell}
\usepackage{tablefootnote}

\usepackage{xspace}
\makeatletter
\DeclareRobustCommand\onedot{\futurelet\@let@token\@onedot}
\def\@onedot{\ifx\@let@token.\else.\null\fi\xspace}
\def\eg{\emph{e.g}\onedot}

\def\ie{\emph{i.e}\onedot}

\makeatother

\title{\LARGE \bf
Training of Convolutional Networks on Multiple Heterogeneous Datasets for Street Scene Semantic Segmentation}

\author{Panagiotis Meletis$^{}$ and Gijs Dubbelman$^{}$
\thanks{$^{}$Panagiotis Meletis ({\tt\small p.c.meletis@tue.nl}) and Gijs Dubbelman ({\tt\small g.dubbelman@tue.nl}) are with the Department of Electrical Engineering, Eindhoven University of Technology,
        Eindhoven, The Netherlands.
        This project has received funding from the European Union's Horizon 2020 research and innovation programme under grant agreement No 688099.}%
}

\begin{document}

\maketitle
\thispagestyle{empty}
\pagestyle{empty}

\begin{abstract}
We propose a convolutional network with hierarchical classifiers for per-pixel semantic segmentation, which is able to be trained on multiple, heterogeneous datasets and exploit their semantic hierarchy.
Our network is the first to be simultaneously trained on three different datasets from the intelligent vehicles domain, \ie Cityscapes, GTSDB and Mapillary Vistas, and is able to handle different semantic level-of-detail, class imbalances, and different annotation types, \ie dense per-pixel and sparse bounding-box labels.
We assess our hierarchical approach, by comparing against flat, non-hierarchical classifiers and we show improvements in mean pixel accuracy of 13.0\% for Cityscapes classes and 2.4\% for Vistas classes and 32.3\% for GTSDB classes. Our implementation achieves inference rates of 17 fps at a resolution of 520 x 706 for 108 classes running on a GPU.
\end{abstract}

\section{INTRODUCTION}
\label{sec:intro}
Semantic classification is a key task in the perception sub-system of an autonomously driving vehicle ~\cite{janai17computer}. The segmentation task, posed as per-pixel classification, has seen great progress in the past years~\cite{zhao2017survey} due to deep learning techniques. However, two critical challenges that still need to be addressed are: 1) to utilize as much and diverse training data as possible, and 2) to increase the number of recognizable classes from a few dozens to virtually anything that a scene can contain.

In this work, we take steps towards solving both challenges and we present a method that leverages multiple \textit{heterogeneous} datasets, \ie datasets with different classes and annotation types, to train a fully convolutional network for per-pixel semantic segmentation. This approach facilitates better use of available datasets, thereby reducing annotation effort, and increases the number of classes that can be recognized. The datasets that we are using in the context of Highly Automated Driving (HAD) are Cityscapes \cite{c4}, Mapillary Vistas \cite{neuhold2017mapillary}, and GTSDB \cite{houben2013detection}.

\begin{figure}[tb]
	\begin{center}
		\includegraphics[width=1.0\linewidth,trim={6.5cm 19.3cm 6.5cm 0.0cm},clip]{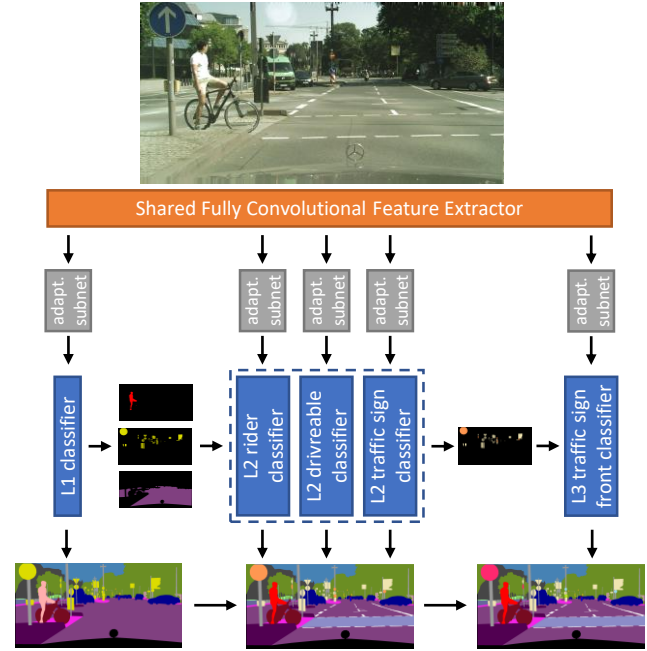}
	\end{center}
	\caption{Our hierarchical classification convolutional network during inference. The input image is transformed to a shared feature representation, which is connected to a hierarchy of classifiers though adaptation subnetworks. The Level-1 classifier outputs predictions for every pixel of the image, while each subsequent classifier infers only about its own set of classes. The output of all levels is combined to form the final fine-grained per-pixel segmentation.}
	\label{fig:algorithm-overview}
\end{figure}

The first challenge, \ie training for semantic segmentation with diverse annotations, is tackled in previous works~\cite{ye2018diverse},~\cite{papandreou2015weakly} by external components to the network, in order to generate pseudo per-pixel ground truth. Our method, in contrast, is self-inclusive and uses the networks' own outputs to refine non-compatible, diverse annotations for supervision.

The second challenge, \ie to increase the number of recognizable classes, can be accomplished in two ways: 1) continue per-pixel annotating an existing dataset with the extra (sub)classes, e.g.~\cite{petrovai2017semi}, or 2) use existing auxiliary datasets only for the new (sub)classes. The first approach can be very costly for big datasets and mainly unnecessary, as a plethora of datasets with fine-grained (sub)classes exist (\eg traffic sign types, car models, pedestrians). In our work, we research the second approach. For this, the heterogeneity, \ie different label spaces and annotations types, of datasets poses challenges for combining them with traditional ``flat'', \ie non-hierarchical, classifiers. Therefore, we propose the use of hierarchical classifiers, which explicitly take advantage of the semantic relationships between the datasets, and we compare against flat classifiers. Our hierarchy is comparable to~\cite{zhou2016hierarchical},~\cite{mao2016hierarchical}, but differs in the scalability it offers.

In Sec.~\ref{sec:challenges}, we describe the exact challenges that are addressed by our hierarchical approach. An example is combined training on Cityscapes and GTSDB. In that case, all classes of GTSDB are subclasses of the traffic sign class in Cityscapes. The straightforward approach of combining classes from both datasets in a conventional flat classifier is infeasible, since a traffic sign pixel cannot have different labels depending on the dataset it comes from. This poses challenges for end-to-end training and inference for flat classifiers, which are addressed by our hierarchical classification scheme.

The fundamentals of our general hierarchical approach are provided in Sec.~\ref{sec:proposed-method} and in Sec.~\ref{sec:three-level-hierarchy} we provide the specifics of our implementation. In Sec.~\ref{sec:evaluation}, we demonstrate the performance gain of hierarchical classifiers with the three \textit{heterogeneous} datasets, over flat, non-hierarchical classifiers. Furthermore, we show that multi-dataset training of a common feature representation, using our proposed method, can improve performance across all datasets regardless of their structural differences.

To summarize, the contributions of this work to per-pixel semantic segmentation are:
\begin{itemize}
	\item A methodology for combined training on datasets with disjoint, but semantically connected, label spaces.
	\item A modular architecture of hierarchical classifiers that can replace the classification stage in modern convolutional networks.
\end{itemize}

Our system implementation is made available to the research community~\cite{panos2017code}. Additionally, we provide our per-pixel annotations of the Cityscapes dataset for the GTSDB traffic sign subclasses, which we use for validation purposes, but is not required for training. We refer to this dataset as Cityscapes Extended throughout this paper.

\section{Challenges from multiple dataset training}
\label{sec:challenges}
End-to-end supervised training on multiple datasets can face many challenges due to the structural differences of the datasets. The most important challenges can be categorized in the following groups:

\subsubsection*{\textbf{Semantic level-of-detail}}
\label{sec:semantic-level-of-detail}
Each dataset is labeled with a set of semantic classes. In the naive flat classification approach, the output of the classifier will be the union of classes from all datasets. It is highly probable that the semantics of a class in one dataset contain, or are contained in, the semantics of a class of another dataset. If those classes are placed in the same level, as in the case of a flat classifier, a conflict of supervision happens, since some pixels belonging to the same semantic class will be labeled with different classes.

For our three datasets, this challenge appears in three cases: 1) Cityscapes defines its road class as: ``part of ground on which cars usually drive'', which includes lane markings, bicycle lanes, potholes, etc. In Vistas, these fine-grained subclasses are separately labeled, in addition to the a road class, leading to conflicts in the semantic level-of-detail of labels, 2) Cityscapes and Vistas contain one high-level traffic sign class, while GTSDB has 43 traffic sign subclasses, and 3) Cityscapes has only one rider class, while Vistas differentiates between three different rider subclasses. Introducing a hierarchy of labels, as shown in Fig.~\ref{fig:label-tree}, effectively solves this challenge, and is discussed in more detail in Sec.~\ref{subsec:semantic-hierarchy-of-label-spaces}.

\subsubsection*{\textbf{Annotation types}}
\label{sec:annotation-types}
Semantic segmentation is by definition a per-pixel problem, thus supervision must be provided at the pixel level. Unfortunately, many existing datasets have bounding-box or per-image annotations, which are incompatible for per-pixel training. The direct approach to make them compatible would be to convert those annotations to masks. However, these masks will include pixels that don't belong to the object of interest, for example bounding boxes might include many non-relevant background pixels that will be assigned to the foreground class. Eventually, during training, supervision to the network will flow from incorrectly labeled pixels, leading to weights confusion.

In our case, Cityscapes and Vistas have per-pixel annotations, but GTSDB has only bounding box annotations. In order to include GTSDB during training, we propose a novel hierarchical loss, provided in~\ref{subsec:training}, which uniformly handles supervision from different annotation types.

\subsubsection*{\textbf{Training sample imbalances}}
\label{sec:sample-imbalances}
Batch-wise training suffers from class imbalances, especially when there are limited examples per batch. In our case, we face strong intra-dataset and inter-dataset imbalances. Imbalances between datasets happen due to the big different of annotated pixels, \ie in the order of $ 10^3 $ (see Table~\ref{tab:datasets}). Imbalances between the classes of the same dataset are common for street scene datasets, since most of the pixels belong to classes of big surfaces, like road and buildings. Our method deals with imbalances by placing classes with the similar order of examples in the same classifier and thus all classes have bigger probability to be represented in the same batch. This strategy is highly beneficial, as shown in Sec.~\ref{subsec:hierarchical-vs-flat}.

\begin{figure*}
	\centering
	\includegraphics[width=1.0\linewidth]{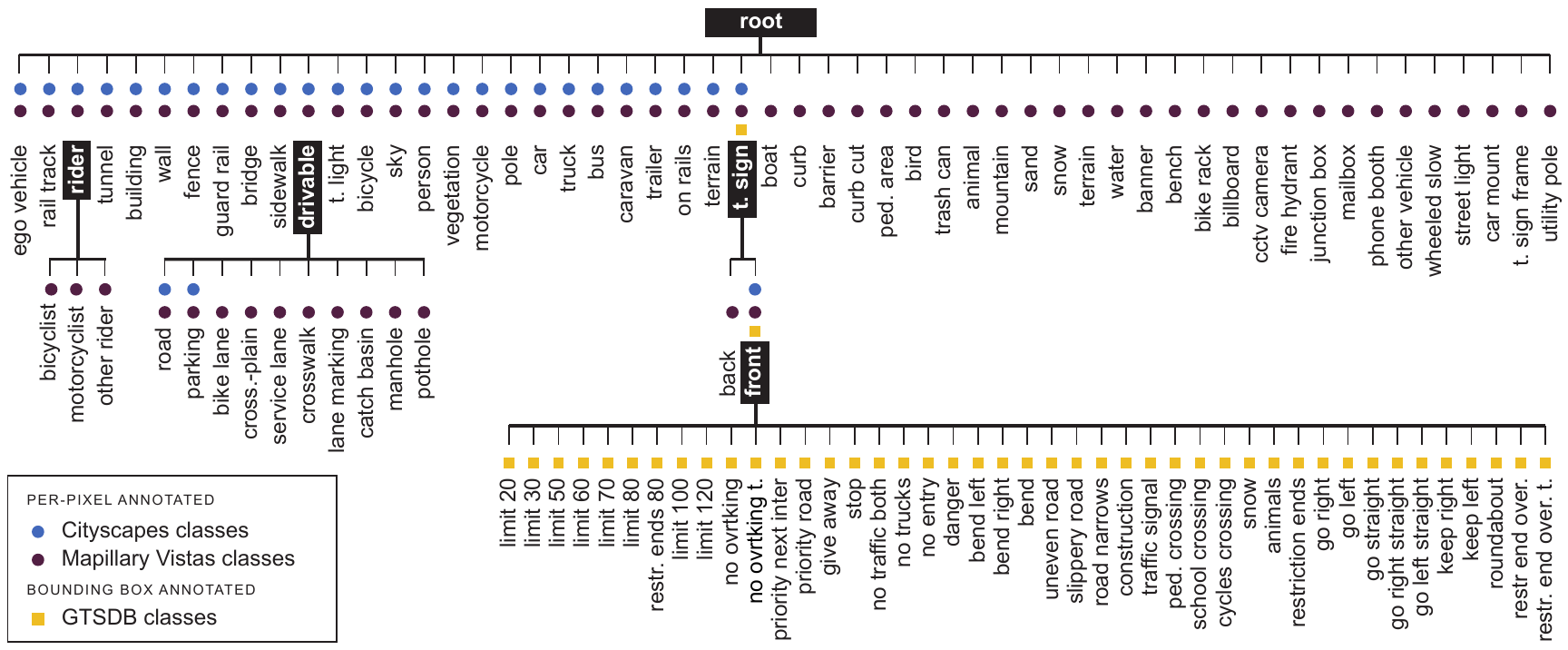}
	\caption{Three-level semantic label hierarchy combining 108 classes from Cityscapes, Mapillary Vistas and GTSDB dataset. Classes marked in black correspond to the L1, L2, and L3 classifiers of Fig.~\ref{fig:algorithm-overview}.}
	\label{fig:label-tree}
\end{figure*}

\section{Training and inference on heterogeneous datasets using semantic hierarchies}
\label{sec:proposed-method}
In this Section, we describe the components of our general hierarchical classification methodology for an arbitrary number of \textit{heterogeneous} datasets. These components provide solutions to the challenges of Sec.~\ref{sec:challenges} and for each one we elaborate on the specific cases for our selected datasets. Our current experiments, detailed in Sec.~\ref{sec:evaluation}, are based on an implementation with a 3-level hierarchy using 3 datasets. The specifics of this implementation are provided in Sec.~\ref{sec:three-level-hierarchy}.

\subsection{Semantic hierarchy of label spaces}
\label{subsec:semantic-hierarchy-of-label-spaces}
Multiple dataset training requires a common label space for all selected datasets. We propose to combine individual label spaces into the common space, containing labels from all datasets, by a hierarchical manner into a semantic tree of labels. This approach solves any conflict in the semantic definition of labels, by introducing the necessary parent or intermediate nodes and/or grouping of existing labels.

Figure~\ref{fig:label-tree}, depicts the 3-level hierarchy using all labels for the three selected datasets of this paper. The challenges that arise from combining these three label spaces were described in Sec.~\ref{sec:semantic-level-of-detail} and are solved as follows: 1) A new high-level driveable class is introduced to solve Cityscapes and Vistas road class semantic conflict, 2) A superclass of traffic signs and an intermediate node for differentiating Vistas \textit{back} from \textit{front} traffic signs are added, and 3) A rider superclass is introduced to include the Cityscapes rider class and the 3 Vistas rider subclasses.

The semantic hierarchy of the labels induces a corresponding hierarchy of classifiers. Each classifier classifies the children labels of a node and the whole tree of classifiers is trained, in an end-to-end, fully convolutional manner, over a shared feature representation.

\subsection{Convolutional network architecture}
\label{subsec:conv-netw-arch}
The proposed network architecture (see Fig.~\ref{fig:algorithm-overview} for an example) consists of a fully convolutional feature extractor for computing a dense, shared representation, and a set of classifiers, each corresponding to an inner class node of the semantic hierarchy. Every classifier can be connected with classifiers one level down in the hierarchy, in order to pass its predictions for inference, and annotation type independent training, as described in Sec. \ref{subsec:inference}, \ref{subsec:training}. Each classifier may be preceded by a shallow \textit{adaptation network}, which adapts the common representation, its depth, and receptive field to the needs of the classifier. This gives the network designer the opportunity to select different features dimensions and receptive fields for each of the classifiers. For example, discriminating between \eg traffic signs is easier~\cite{ciregan2012multi}, as less features are needed, compared to high-level discrimination, like road vs. sidewalk and bushes vs. trees~\cite{c4}. The flexibility of applying a different field-of-views to different classifiers, enables more or less context aggregation, depending on the classifier's object average size: \eg traffic signs appear generally in smaller scales than buildings or cars.

\subsection{Inference: hierarchical decision rule}
\label{subsec:inference}
Inference is carried out per-pixel, in a hierarchical manner across the tree of softmax classifiers. Each classifier $ j $ computes a per-pixel normalized vector $ \bsigma^{j,p} $ of class probabilities for its own set of pixels $ p \in P^{j} $ and set of classes $ C^{j} = \{0, 1, ...\} $, and outputs per-pixel decisions $ \hat{y}^{j,p} = \argmax_i \bsigma_i^{j,p} $, where $ \hat{y}^{j,p} \in C^j $. The set of pixels $ P^{j} $, for which every classifier must produce decisions, is generated by its parent according to its own decisions. Every pixel of the input is labeled with the desired level of detail, from the available set of labels $ \{\hat{y}^{j,p}\}_{j \in J} $, where $ J $ are the classifiers that produced decisions for this specific pixel.

\subsection{Training: hierarchical classification loss}
\label{subsec:training}
The annotations type of many datasets are not compatible with the required per-pixel supervision for semantic segmentation, as outlined in Sec.~\ref{sec:annotation-types}. Our proposed method treats incompatible annotations with a unified approach, without the need of external components, as in~\cite{ye2018diverse},~\cite{papandreou2015weakly}, and with negligible computational load to the system. The flexibility for handling diverse ground truth, is exchanged with the only constraint that classes on the root classifier, should have per-pixel annotated examples. Annotations for any other level can be of any type or even mixed.

\begin{figure}
	\begin{center}
		\includegraphics[width=1.0\linewidth,trim={0.0cm 16.0cm 26.5cm 0.0cm},clip]{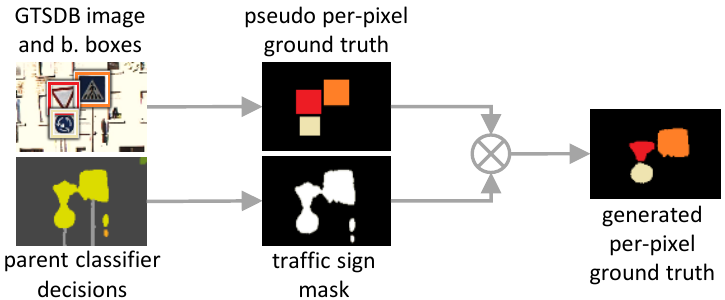}
	\end{center}
	\caption{Online procedure during training for generating per-pixel ground truth from bounding box labels.}
	\label{fig:semantic-masks}
\end{figure}

We propose a hierarchical classification loss, which separates supervision according to annotations type at the pixel level. Each classifier $ j $ is trained on all labeled pixels $ P^j = P_1^j + P_2^j $ that correspond to its respective node in the label hierarchy. Pixels $ P_1^j $ with per-pixel annotations are trained using the standard one-hot cross-entropy loss. Pixels $ P_2^j $ with non-per-pixel annotations are trained with generated per-pixel ground truth using a modified cross-entropy loss. 
To achieve this, our method uses the online, per-pixel decisions of the parent classifier during training, to refine the pseudo, per-pixel labels.
The process is illustrated in Fig.~\ref{fig:semantic-masks}. First, non-compatible annotations are converted to per-pixel pseudo ground truth.
Then, in every training step, decisions of parent classifiers are intersected with this pseudo ground truth to generate the per-pixel ground truth used for supervision.

Both losses are accumulated per classifier to the so called hierarchical loss:
\begin{equation}
\label{eq:ss-losses}
L^j = - \frac{1}{\abs*{P_1^j}} \sum_{p \in P_1^j} \log \bsigma^{j,p}_{y^{j,p}} - \frac{1}{\abs*{P_2^j}} \sum_{p \in P_2^j} \log \bsigma^{j,p}_{y^{j,p}} ~,
\end{equation}
where $ \abs*{\cdot} $ is the cardinality of the pixel's set, and $ y^{j,p} \in C^j $ selects the element of $ \bsigma $ that corresponds to the ground truth class of pixel $ p $ for classifier $ j $. Finally, losses from all classifiers are collected and weighted with different hyperparameters $ \lambda^j $ to obtain the total objective to minimize:
\begin{equation}
\label{eq:total-loss}
L^{total} = \sum_j \lambda^j \cdot L^j + regularizer ~.
\end{equation}

\section{Three-level label hierarchy with Cityscapes, Mapillary Vistas and GTSDB}
\label{sec:three-level-hierarchy}
In this section, we outline implementation details, to improve the repeatability of our experiments.

\subsubsection*{\textbf{Convolutional network architecture}}
The network is depicted in Fig.~\ref{fig:algorithm-overview}. The feature extractor consists of the feature layers of the ResNet-50 architecture~\cite{he2016deep}, followed by an 1x1 convolutional layer (with ReLU and Batch Normalization), to decrease feature dimensions to 256. The stride on the input is reduced from 32 to 8, using dilated convolutions. The representation has depth 256, spatial dimensions 1/8 of the input, and is shared among 5 branches. Each branch has an extra bottleneck module~\cite{he2016deep} and ends at a softmax classifier, which includes a hybrid upsampling module. We choose the feature dimensions and the field-of-view of the per-classifier adaptation subnetworks to be the same for all branches. After experimenting with different upsampling techniques (fractional strided convolution, bilinear, convolutional), we concluded that the best performance and reduction of artifacts, is obtained by hybrid upsampling, which consists of one 2x2 learnable fractional strided convolutional layer, followed by bilinear upsampling to reach input dimensions.

\subsubsection*{\textbf{Implementation details}}
We use Tensorflow~\cite{abadi2016tensorflow} and a Titan X (Pascal architecture) GPU for training and inference. Due to limited memory we set batch size to 4 (Cityscapes:Vistas:GTSDB = 1:2:1) and training dimensions to 512x706 (the average of Vistas images scaled to the smaller Cityscapes dimension). During training images are downscaled, respecting their aspect ratio, and then randomly cropped. The network is trained for 17 Vistas epochs (early stopping) with Stochastic Gradient Descent and momentum of 0.9, L2 weight regularization with decay of 0.00017, initial learning rate 0.01 that is halved three times, and batch normalization and exponential moving averages decay are both set to 0.9. The hyperparameters $ \lambda^j $ of Eq. (\ref{eq:total-loss}) are chosen to be 1.0, 0.1 and 0.1 for the three levels of the hierarchy respectively. For inference, we currently achieve a frame rate of 17 fps, \ie 58 milliseconds per frame.

\section{Evaluation}
\label{sec:evaluation}
We conduct the following experiments to evaluate our hierarchical classification approach:
\begin{enumerate}
\item \textbf{Baselines for flat classification}: Sets the baselines of flat classifiers for single and multiple datasets training.
\item \textbf{Hierarchical classification on three heterogeneous datasets}: Demonstrates the benefits of our complete method for combined training on three \textit{heterogeneous} datasets (Cityscapes, GTSDB, Vistas) with disjoint label spaces and different annotation types.
\item \textbf{Hierarchical versus flat classification on Cityscapes Extended}: Validates the effectiveness of our hierarchical approach against extremely imbalanced classes, by isolating it on the per-pixel annotated Cityscapes Extended dataset with a two-level label space.
\end{enumerate}

\subsection{Datasets}
\label{subsec:datasets-generation}
We summarize the datasets that we use in Table~\ref{tab:datasets}. Next we describe the extra annotations needed for our experiments. Please note that these annotations are only used for validation purposes and not for training the networks.

\begin{table}[b]
	\setlength\tabcolsep{4.0pt}
	\caption{Dataset statistics. Images contain training and validation splits. In parenthesis the number of evaluated classes are shown.}
	\label{tab:datasets}
	\begin{center}
		\begin{tabular}{c|c|c|c|c}
			& Cityscapes & Cityscapes & Mapillary & \multirow{2}{*}{GTSDB}\\
			& (fine) & Extended & Vistas &\\
			\hline
			resolution & \multicolumn{2}{c|}{1024 x 2048} & 0.5 - 25 MP & 800 x 1360\\
			\hline
			images & \multicolumn{2}{c|}{2975 / 500} & 18000 / 2000 & 600 / 300\\
			\hline
			annot. type & \multicolumn{2}{c|}{per-pixel} & per-pixel & bound. box\\
			\hline
			annot. pixels & \multicolumn{2}{c|}{$1.6 \cdot 10^9$} & {$156.2 \cdot 10^9$} & $0.003 \cdot 10^9$\\
			\hline
			classes & \multicolumn{2}{c|}{34 (27)} & 66 (65) & 43 (28) \\
			\hline
			t. sign classes & - & 43 (18) & - & 43 (28)\\
			\hline
			traffic signs & - & 3158 & - & 900\\
		\end{tabular}
	\end{center}
\end{table}

\subsubsection{Labeling Cityscapes with traffic sign classes}
We extend the label space of Cityscapes with 43 traffic sign classes of GTSDB. Cityscapes provides only per-pixel traffic sign annotations without differentiating between instances. We design an automatic segmentation algorithm based on the 8-neighborhood distance, for separating connected traffic sign instances in the ground truth traffic sign mask, and a GUI application, which proposes image areas for labeling. We pack original and new annotations under the name Cityscapes Extended. This dataset contains 2778 and 380 traffic signs in the train and validation splits respectively.

\subsubsection{Annotating GTSDB with per-pixel labels}
\label{subsec:gtsdb-bboxes-to-perpixel}
Only for specific experiments that involve the flat classifier, we converted the GTSDB bounding box annotations to fine per-pixel annotations, using the traffic signs shapes (circle, triangle, hexagon) and inscribing them into their bounding box. This procedure can be problematic with in-plane rotation of traffic signs, but after dataset inspection, we observed that only a negligible amount of in-plane rotations are present.

\subsection{Metrics and evaluation conventions}
\label{subsec:metrics}
We use multi-class mean Pixel Accuracy (mPA) and mean Intersection over Union (mIoU), which are relevant in the context of automated driving, and they represent good local and area criteria, following the definitions given in~\cite{garcia2017review}. For Cityscapes, we report results on 27 classes (19 of the official benchmark + 8 common with Vistas). For the traffic sign classes we evaluate on a subset of the 43 traffic signs that satisfy both conditions: 1) have less than 10\textsuperscript{3} pixels in GTSDB train set, and 2) have less than 10\textsuperscript{3} pixels on both GTSDB and Cityscapes Extended validation sets. Please note that 10\textsuperscript{3} pixels limit is chosen, as it is 2 orders of magnitude smaller than the least represented class in Cityscapes. For Vistas, we report results on the official 65 classes benchmark. Finally, we evaluate the performance of the model every one epoch and we report the average over the last 2 runs.

A new evaluation protocol for fair comparisons is introduced only for the experiments of Sec.~\ref{sec:baselines}, which trains a flat classifier on two datasets. It solves the semantic conflict of the high-level traffic sign class being in the same level with traffic sign subclasses (Sec.~\ref{sec:semantic-level-of-detail}). The decision for a traffic sign pixel is deemed correct: 1) if it is correctly labeled with any traffic sign subclass, or 2) if it is labeled as traffic sign and the second most probable choice is the correct traffic sign subclass. To be clear, we do not use this evaluation scheme for the hierarchical classifier but only for the flat classifier.

\subsection{Baselines for flat classification}
\label{sec:baselines}
In Table~\ref{tab:flat-baselines}, we set the same- and cross-dataset baselines for the conventional flat classification approach, using the same input dimensions and batch size as described in the implementation details of Sec.~\ref{sec:three-level-hierarchy}, in order to be able to fairly compare with the hierarchical results of Table~\ref{tab:hierarchical}. In columns 1-3, we train three models on three datasets independently and we provide results for the evaluated classes of Table~\ref{tab:datasets}. In column 4, we provide cross-dataset results on Cityscapes Extended of combined training on Cityscapes and GTSDB.

For fair comparisons, the models of third and fourth column were trained with the generated per-pixel annotations of the GTSDB dataset (see Sec.~\ref{subsec:gtsdb-bboxes-to-perpixel} for the details). Training on 43 classes of GTSDB does not converge, due to the limited number of training pixels per image, so we included the unlabeled pixels as an extra class, to solve this issue. It can be observed that simultaneous training on Cityscapes and GTSDB, fails to achieve satisfactory cross-dataset results on the traffic sign classes of Cityscapes Extended.

\begin{table}
	\setlength\tabcolsep{3.5pt}
	\caption{Flat classification performance baselines on per-pixel annotated datasets.}
	\label{tab:flat-baselines}
	\begin{center}
		\begin{tabular}{c||c|c|ccc}
			& \multicolumn{3}{c}{Same dataset} & & Cross-dataset\\
			\cline{2-4} \cline{6-6}
			\multirow{2}{*}{Tested on} & \multirow{2}{*}{Cityscapes} & \multirow{2}{*}{Vistas} & \multirow{2}{*}{GTSDB}  & & Cityscapes Extended\\
			& & & & & traffic sign classes\\
			\cline{1-4} \cline{6-6}
			mPA (\%) & 53.6 & 36.5 & 25.4 & & 19.1\\
			mIoU (\%) & 46.2 & 29.6 & 17.2 & & 3.0\\
			\cline{1-4} \cline{6-6}
			Trained on & Cityscapes & Vistas & GTSDB & & Cityscapes + GTSDB
		\end{tabular}
	\end{center}
\end{table}

\subsection{Hierarchical classification on 3 heterogeneous datasets}
This experiment evaluates our complete hierarchical classification approach on three \textit{heterogeneous} datasets (Cityscapes, Mapillary Vistas, and GTSDB).

\begin{figure}
	\centering
	\includegraphics[width=1.0\linewidth,trim={4.4cm 21.5cm 4.2cm 0.0cm},clip]{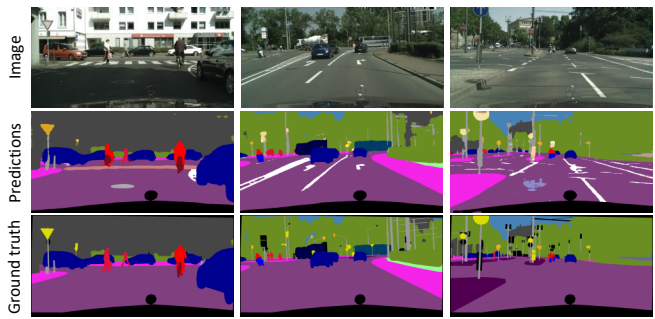}
	\caption{Cityscapes val split image examples. The network predictions include decisions from L1-L3 levels of the hierarchy. Note that the ground truth includes only one traffic sign superclass (yellow) and no road attribute markings.}
	\label{fig:results-cityscapes}
\end{figure}

\begin{figure}
	\centering
	\includegraphics[width=1.0\linewidth,trim={4.4cm 18.5cm 4.2cm 0.0cm},clip]{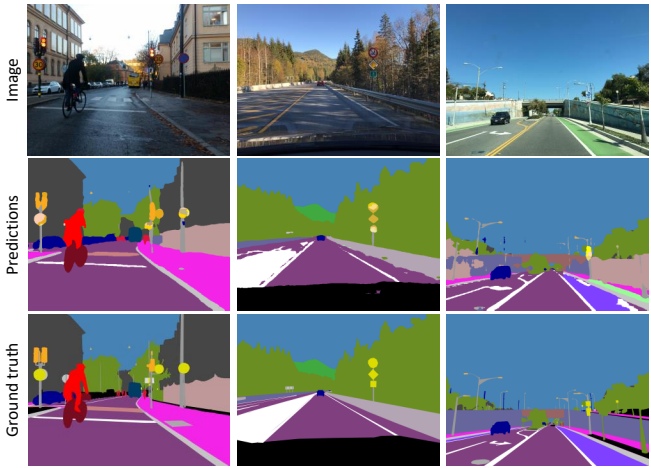}
	\caption{Mapillary Vistas validation split image examples. The network predictions include decisions from L1-L3 levels of the hierarchy. Note that the ground truth does not include traffic sign subclasses.}
	\label{fig:results-vistas}
\end{figure}

\begin{figure}
	\centering
	\includegraphics[width=1.0\linewidth,trim={4.4cm 20.5cm 4.2cm 0.0cm},clip]{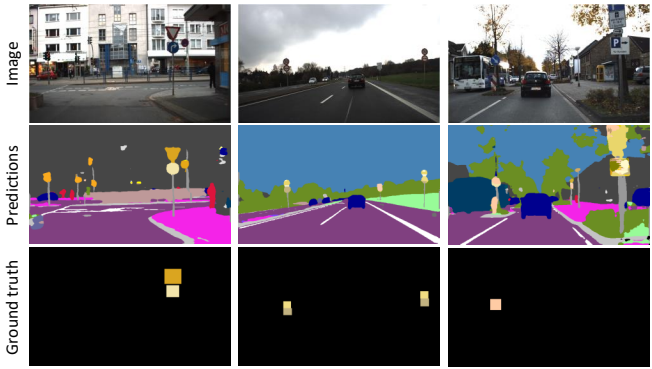}
	\caption{GTSDB test split image examples. The network predictions include decisions from L1-L3 levels of the hierarchy. Note that the ground truth includes only traffic sign bounding boxes, since rest pixels are unlabeled.}
	\label{fig:results-gtsdb}
\end{figure}

In Table~\ref{tab:hierarchical}, we present evaluation results on the validation splits of the three datasets that the model is trained on (columns 1-3) and results for traffic sign subclasses on Cityscapes Extended (column 4), which was not used during training. In Figures~\ref{fig:results-cityscapes},~\ref{fig:results-vistas},~\ref{fig:results-gtsdb} qualitative results are depicted.

\begin{table}
	\setlength\tabcolsep{3.5pt}
	\caption{Performance of our complete hierarchical classification approach on 4 datasets.}
	\label{tab:hierarchical}
	\begin{center}
		\begin{tabular}{c||c|c|ccc}
			& \multicolumn{3}{c}{Same dataset} & & Cross-dataset\\
			\cline{2-4} \cline{6-6}
			\multirow{2}{*}{Tested on} & Cityscapes & Vistas & GTSDB & & {Cityscapes Extended}\\
			& classes & classes & classes & & traffic sign classes\\
			\cline{1-4} \cline{6-6}
			mPA (\%) & 66.6 & 38.9 & 57.7 & & 29.7\\
			mIoU (\%) & 57.3 & 31.9 & 41.5 & & 8.3\\
			\cline{1-6}
			Trained on & \multicolumn{5}{c}{Cityscapes + Vistas + GTSDB}
		\end{tabular}
	\end{center}
\end{table}

By comparing Table~\ref{tab:flat-baselines} columns 1-3 and Table~\ref{tab:hierarchical} columns 1-3, we achieve significant performance increases in mean PA (in the range +2.4\% to +32.3\%) and IoU (in the range +2.3\% to +24.3\%) for all datasets. By comparing Table~\ref{tab:flat-baselines} column 4 and Table~\ref{tab:hierarchical} column 4, we also observe an increase in cross-dataset performance on traffic sign subclasses. It is important to note that the model was not trained on any examples from Cityscapes Extended traffic sign classes, and the +10.6\% increase in mean PA is solely due to our hierarchical multiple dataset training scheme.

We conclude that hierarchical classification is highly advantageous for combined \textit{heterogeneous} datasets training, when datasets have different classes, different annotation types and in- and between-dataset(s) imbalances.

\subsection{Hierarchical vs flat classification on Cityscapes Extended}
\label{subsec:hierarchical-vs-flat}
In this experiment we evaluate the hierarchical classification method on Cityscapes Extended with per-pixel annotations and a two-level label space. The goal is to isolate our method on a single dataset, in order to show its effectiveness on highly imbalanced datasets against flat classification. We used 512 x 1024 input dimensions and batch size of 2.

From Table~\ref{tab:flat-vs-hierarchical}, we observe that hierarchical classification significantly increases mPA (+26.0\%) and
mIoU (+16.1\%) for L2 classes (i.e. GTSDB traffic sign subclasses) with respect to the flat classifier, while for L1 classes (i.e. Cityscapes classes) the increase is above +6\% for both mPA and IoU.

We conclude that hierarchical classification is robust against class imbalances, even when applied on a single dataset with per-pixel annotations, since it places in each level classes with the same order of examples.

\begin{table}
	\centering
	\caption{Flat versus proposed hierarchical classification performance on Cityscapes Extended. (In parenthesis performance for traffic sign l1 class.)}
	\begin{tabular}{c||c|c||c|c}
		& \multicolumn{2}{c||}{flat classifier} & \multicolumn{2}{c}{hierarchical classifiers}\\
		& L1 classes & L2 classes & L1 classes & L2 classes\\
		\hline
		mPA (\%) & 69.4 (73.0) & \textbf{23.0} & 75.6 (74.8) & \textbf{49.0}\\
		mIoU (\%) & 60.4 (65.2) & \textbf{12.7} & 66.7 (65.7) & \textbf{28.8}\\
		\hline
		trained on & \multicolumn{4}{c}{Cityscapes Extended L1 and L2 classes}
	\end{tabular}
	\label{tab:flat-vs-hierarchical}
\end{table}

\section{CONCLUSIONS AND FUTURE WORK}
In this paper, we considered the challenge of simultaneously training on three \textit{heterogeneous}, but semantically connected datasets, for the problem of per-pixel semantic segmentation. The main motives are to maximally reuse resources (datasets and computation) and eliminate human labeling effort. In order to achieve this, we leverage the semantic relationships between datasets' labels to construct a hierarchy of classifiers and we introduce the respective hierarchical training and inference rules. Our final network can per-pixel segment an input image into 108 classes from 8 high-level street scene categories. The results clearly show the benefit of using our hierarchical classification scheme for multiple heterogeneous dataset training. In future work, we will extend our implementation to include more datasets with more divergent characteristics, to demonstrate the scalability of our approach.


\bibliographystyle{IEEEtran}
\bibliography{IEEEabrv,biblio}

\end{document}